\pdfoutput=1
\documentclass[11pt]{article}
\usepackage{amsmath}

\usepackage{multirow}
\usepackage{tabularx, booktabs}
\usepackage[table]{xcolor}
\newcolumntype{?}{!{\vrule width 1pt}}
\newcolumntype{|}{!{\vrule width .5pt}}
\usepackage{arydshln}
\usepackage{float}
\usepackage[final]{acl} %review

\usepackage{times}
\usepackage{latexsym}

\usepackage[T1]{fontenc}
\usepackage[utf8]{inputenc}

\usepackage{microtype}

\usepackage{inconsolata}

\usepackage{graphicx}

\title{Ask Patients with Patience: Enabling LLMs for Human-Centric Medical Dialogue with Grounded Reasoning}

\author{Jiayuan Zhu \\ University of Oxford
  \And
  Jiazhen Pan \thanks{Corresponding author} \\ Technical University of Munich
    \\\And
  Yuyuan Liu \\ University of Oxford
    \\\AND
  Fenglin Liu \\ University of Oxford
  \\\And
  Junde Wu \\ University of Oxford
  }

\begin{document}
\maketitle
\begin{abstract}
    The severe shortage of medical doctors limits access to timely and reliable healthcare, leaving millions underserved. Large language models (LLMs) offer a potential solution but struggle in real-world clinical interactions. Many LLMs are not grounded in authoritative medical guidelines and fail to transparently manage diagnostic uncertainty. Their language is often rigid and mechanical, lacking the human-like qualities essential for patient trust. To address these challenges, we propose \textit{Ask Patients with Patience (APP)}, a multi-turn LLM-based medical assistant designed for grounded reasoning, transparent diagnoses, and human-centric interaction. APP enhances communication by eliciting user symptoms through empathetic dialogue, significantly improving accessibility and user engagement. It also incorporates Bayesian active learning to support transparent and adaptive diagnoses. The framework is built on verified medical guidelines, ensuring clinically grounded and evidence-based reasoning. To evaluate its performance, we develop a new benchmark that simulates realistic medical conversations using patient agents driven by profiles extracted from real-world consultation cases. We compare APP against SOTA one-shot and multi-turn LLM baselines. The results show that APP improves diagnostic accuracy, reduces uncertainty, and enhances user experience. By integrating medical expertise with transparent, human-like interaction, APP bridges the gap between AI-driven medical assistance and real-world clinical practice. 

\end{abstract} \label{abstract}

%%%%%%%%%%%%%%%%%%%%%%%%%%%%%%%%%%%%%%%%%%%%%%%%%%
\section{Introduction} 
The shortage of medical doctors is a critical global issue. It is noteworthy that 40\% of WHO Member States report having fewer than ten medical doctors per 10,000 people, with over 26\% having fewer than three \cite{noauthor_medical_nodate}. Large language models (LLMs), such as the GPT series \cite{radford2018improving, radford2019language, brown2020language, ouyang2022training, achiam2023gpt}, have significantly improved access to medical inquiries. Notably, models such as GPT-4 with Medprompt \cite{nori2023can}, Med-PaLM 2 \cite{singhal2025toward}, and Med-Gemini-L 1.0 \cite{saab2024capabilities} have achieved expert-level performance on benchmarks like MedQA (USMLE) \cite{jin2021disease}, claiming to surpass human experts in structured evaluations. Beyond standard medical question-answering benchmarks, several approaches have demonstrated strong diagnostic capabilities when provided with comprehensive real-world patient cases \cite{kanjee2023accuracy, rios2024evaluation, mcduff2025towards}.

\begin{figure*}[hbt!]
    \centering
    \includegraphics[width=1\linewidth]{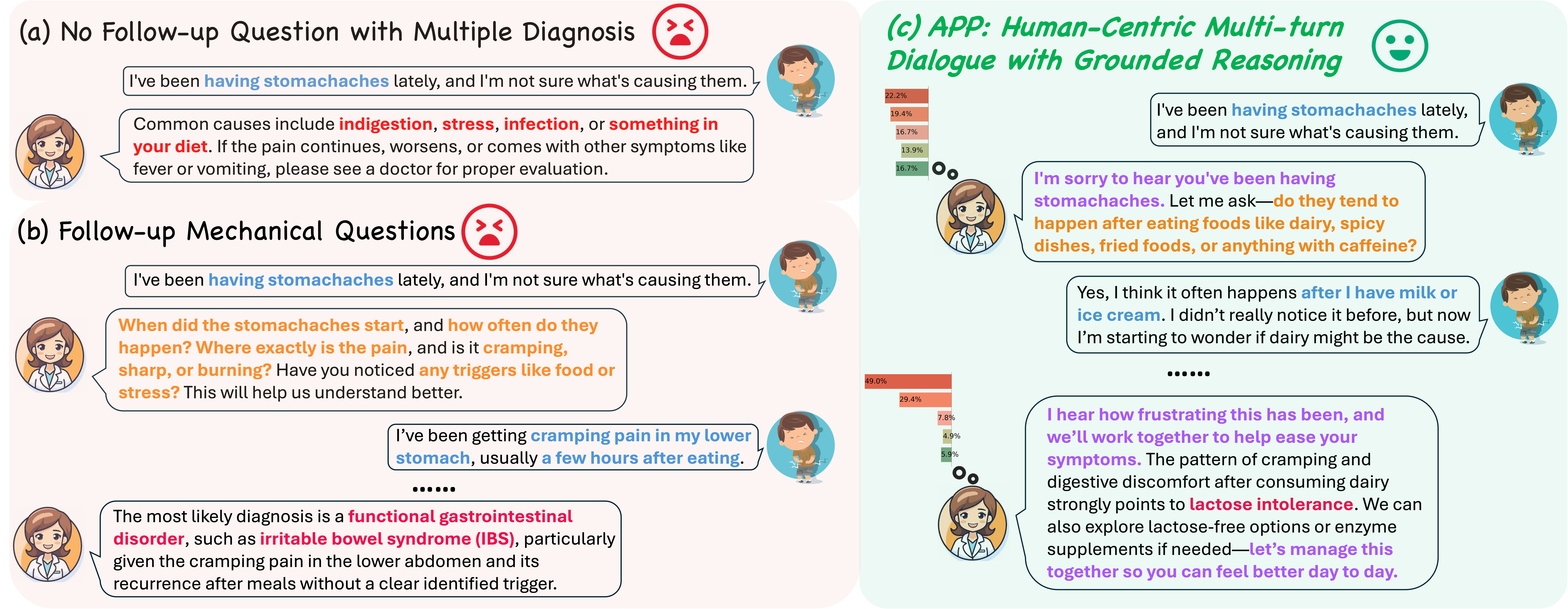} %width=0.75\linewidth
    \caption{(a) Existing LLMs follow a one-shot diagnostic approach, generating multiple possible diseases without asking follow-up questions. (b) While LLMs can be prompted for multi-turn dialogues, they often overwhelm users with excessive mechanical inquiries, potentially disrupting the dialogue and reducing engagement. (c) Our human-centric multi-turn dialogue with grounded reasoning approach, APP, structures follow-up questions in a logical sequence. It incorporates grounded medical sources to build a statistical model, improving reliability and transparency in handling diagnostic uncertainty. It also incorporates human-centric features, such as eliciting patient symptoms with empathy to reduce user pressure and anxiety. \textcolor[RGB]{78, 149, 217}{Blue} represents user-described symptoms, \textcolor[RGB]{255, 133, 21}{Orange} indicates medical assistant questions, \textcolor[RGB]{240, 14, 77}{Red} highlights the diagnosis, and \textcolor[RGB]{184, 82, 255}{Purple} shows human-centric features.}
    \label{fig:overflow}
\end{figure*}

Although current LLMs exhibit expert-level proficiency, they remain difficult to implement in clinical practice. A major limitation is their inability to elicit a patient's most relevant medical conditions through conversational interaction. Notably, most of them identify diseases solely based on the user’s initial input without follow-ups (Fig.\ref{fig:overflow}(a) \cite{wang2023huatuo, xiong2023doctorglm, gupta2025digital}). But in practice, patients often struggle to provide all relevant information in the first place. In contrast, real-world human doctors will have a long conversation with patients, using empathetic questioning to elicit patients' most relevant health concerns. A straightforward approach to LLM-assisted diagnosis is to prompt models to engage in multi-turn dialogues with patients (Fig. \ref{fig:overflow}(b)), which has been shown to be more effective than one-shot consultations \cite{schmidgall2024agentclinic}. However, this approach remains impractical in real-world scenarios due to the following key challenges.

%-------------------------------------------------------------------------
%-------------------------------------------------------------------------

First, LLM-generated language often lacks human-like qualities, making interactions feel mechanical, impersonal, and ineffective, which can even negatively impact diagnosis \cite{xu2019end, bao2023disc, chen2023knse}. In real clinical settings, patients often struggle to accurately describe their symptoms or overlook clinically relevant details. For example, a person with lactose intolerance might only report general stomach discomfort without realizing the link to dairy consumption. A key capability of human doctors is guiding patients toward articulating unrecognized but medically important conditions. Rather than asking broad, generic questions like \textit{``What food might have triggered your symptoms?''}—as LLM-based models might—a doctor might instead ask a more accessible and context-aware question such as \textit{``Did you drink milk last night?''}. This helps patients share clearer and more relevant responses. 

Another major challenge in LLM-based medical consultations is their black-box nature. LLMs may generate hallucinations \cite{xu2024hallucination}, offer inconsistent responses to the same question, use obscure medical terminology without clear sources, and make deterministic medical decisions without grounded reasoning \cite{ness2024medfuzz, ullah2024challenges, shi2024medical, pan2025medvlm}. These limitations undermine transparency and trustworthiness, making it difficult for LLMs to provide reliable diagnoses and gain patient trust, ultimately constraining their real-world clinical applicability.

For LLM-simulated medical assistants to be applied effectively in the real world, they must incorporate human-centric features \cite{busch2025current, lin2025roles}. Using simple, people-friendly language helps patients better understand and respond to medical questions. Guided questioning based on personal background—such as daily activities and dietary habits—can elicit potential health conditions that might otherwise be overlooked. Anthropomorphic features, such as empathetic dialogue, help users feel comfortable and psychologically supported. This improves the user experience and builds trust, fostering a friendly doctor-patient relationship \cite{vishwanath2024role}.

%-------------------------------------------------------------------------
\begin{figure*}[hbt!]
    \centering
    \includegraphics[width=0.98\linewidth]{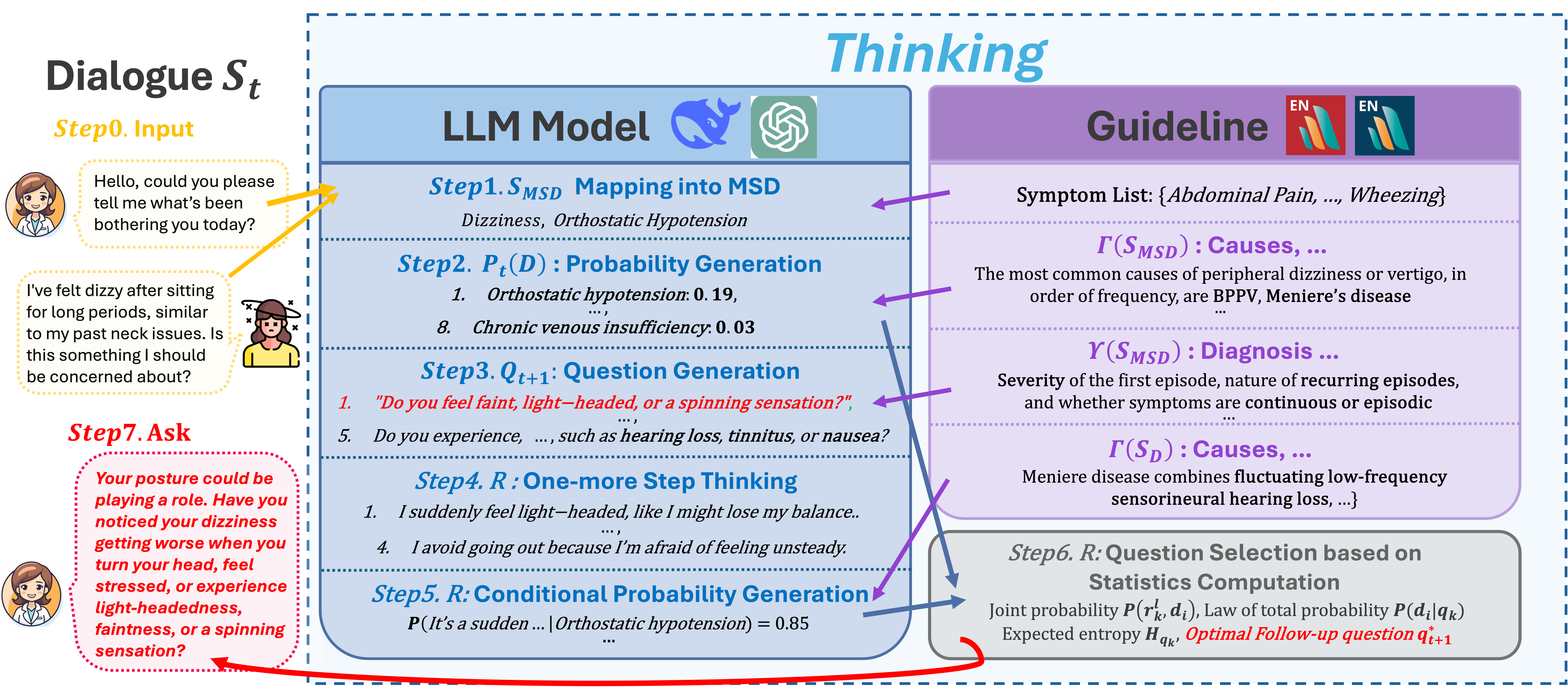} %width=0.75\linewidth
     \caption{APP Workflow. 
     The system first maps dialogue $S_t$ to $S_{MSD}$ symptoms, then generates disease probabilities $P_t(D)$ and a question pool $Q_{t+1}$ based on the MSD Manual \cite{noauthor_msd_nodate-1, noauthor_msd_nodate}. It then performs an additional reasoning step to simulate possible responses, compute conditional probabilities, and apply Bayesian active learning to identify the question with the lowest entropy. This question is then returned to the doctor for further inquiry. \textcolor[RGB]{255, 192, 0}{Yellow} arrows represent the input, \textcolor[RGB]{146, 65, 209}{Purple} arrows indicate the grounded medical guidelines, \textcolor[RGB]{78, 111, 166}{Blue} arrows represent the statistics to compute question selection, and \textcolor[RGB]{240, 14, 77}{Red} highlights the final step, which determines the optimal question to ask. }
    \label{fig:workflow}
\end{figure*}
%-------------------------------------------------------------------------

In this paper, we propose \textbf{A}sk \textbf{P}atients with \textbf{P}atience (\textbf{APP}), a new LLM-based clinical dialogue model designed for grounded reasoning and human-centric interactions. We simulate an anthropomorphic medical assistant, Dr.APP, designed to provide grounded, transparent, and accurate diagnoses. First, Dr.APP strictly follows clinical-standard medical guidelines, ensuring reliable and evidence-based diagnoses. Second, Dr.APP is built on an analytical mathematical model, specifically Bayesian active learning, to determine the optimal next question at each turn. This enhances transparency, improves diagnostic confidence, and maintains high diagnostic accuracy. Finally, Dr.APP facilitates human-centric dialogue by guiding patients to clearly express their symptoms through empathetic communication. Dr.APP is instructed to respond with understanding and compassion, treating user concerns as if in a conversation with a trusted friend. To evaluate our method, we develop a new benchmark that simulates patients using profiles constructed from over 300 real-world doctor interviews \cite{yan2022remedi}. Both medical professionals and non-medical participants assessed the model from complementary perspectives, such as diagnostic accuracy from a clinician's view and empathy from a patient’s view. In summary, our contributions are:

\begin{itemize}
\item We introduce Dr.APP, the first \textbf{human-centric} LLM-based medical assistant capable of eliciting user symptoms through natural and human-like dialogue. It significantly improves user accessibility and engagement.
\item Dr.APP incorporates Bayesian active learning based on authoritative medical guidelines to provide \textbf{grounded and transparent reasoning} for medical diagnosis.
\item We develop a new benchmark that simulates clinical consultations using real-world interview cases. Dr.APP achieves SOTA diagnostic accuracy and provides an \textbf{empathetic user experience}, supported by human evaluation.

\end{itemize}

%%%%%%%%%%%%%%%%%%%%%%%%%%%%%%%%%%%%%%%%

%%%%%%%%%%%%%%%%%%%%%%%%%%%%%%%%%%%%%%%%
% \section{Related Work}

%%%%%%%%%%%%%%%%%%%%%%%%%%%%%%%%%%%%%%%%
\section{Methodology}
\subsection{Framework Overview}
To ensure grounded reasoning and diagnostic reliability, we use authoritative clinical guidelines as the primary source throughout the workflow. Building on this, we incorporate Bayesian active learning to enhance transparency and accuracy. Additionally, we design the medical assistant with a human-centric approach for more empathetic interactions.

The dialogue begins with Dr.APP asking the first question $q_1$ and the patient replying with $r_1$, forming the initial conversation $S_{1}  = {(q_1, r_1)}$. LLMs then extract symotom information from $S_{1}$ and map it to pre-defined symotom list derived from clinical guidelines. Dr.APP aims to identify the most probable diagnosis $d^{*} \in D$, where $D=\{d_i\}_{i=1}^{I}$ represents the set of possible diseases. Let $S_t$ denote the dialogue between the user and Dr.APP after $t$ iterations: $S_t=\{(q_1,r_1), \dots, (q_t, r_t)\}$. At each iteration $t$, the disease probability distribution $P_t(D)$ is updated using $S_t$ and relevant medical knowledge from clinical guidelines. Based on the extracted symptoms, a question pool $Q_{t+1}$ is generated, also guided by the verified clinical guidelines. Dr.APP selects the optimal follow-up question $q_{t+1}^*$ \textit{via} Bayesian active learning for the next iteration.

Specifically, Dr.APP follows the steps shown in Figure \ref{fig:workflow}: Mapping into clinical guidelines (Section \ref{sec:initial_map_msd}); Diagnosis Probability Prediction (Section \ref{sec:potential_diagnosis_probability}); Question Generation (Section \ref{sec:question_generation}); One-more Step Thinking \& Conditional Probability Generation (Section \ref{sec:one_more_think}); Question Selection (Section \ref{sec:question_selection}); and Human-centric Communication (Section~\ref{sec:human_friendly}).

\subsection{Mapping into MSD}
\label{sec:initial_map_msd}
In this work, we incorporate both the professional and consumer versions of the MSD Manual \cite{noauthor_msd_nodate-1, noauthor_msd_nodate} as sources of clinical guidance, though other reliable guidelines can also be used. The professional version offers structured definitions and diagnostic criteria, while the consumer version provides simplified explanations accessible to general users. This dual use ensures that Dr.APP remains medically grounded while interpretable for non-expert users. To map user dialogue to MSD information, we store symptom-description pairs locally and apply RAG to retrieve the most relevant symptoms. Specifically, Dr.APP ensures a comprehensive representation by mapping the initial dialogue $S_1$ to symptoms in both the professional and consumer symptom lists of the MSD Manual: $S_{MSD} = \{S_{prof}, S_{cons}\}$.

\subsection{Diagnosis Probability Prediction}
\label{sec:potential_diagnosis_probability}
Given the MSD symptom set $S_{MSD}$, we access the detailed symptom page, which provides information on the \textit{causes}, \textit{pathophysiology}, and \textit{etiology} of the symptom. We retrieve these sets of information and represent them as $\Gamma(S_{MSD})$. This reliable medical knowledge, combined with the current dialogue context $S_t$, serves as the foundation for generating the potential disease probability distribution:
\vspace{-15pt}
\begin{multline}
P_t(D \mid \Gamma(S_{MSD}), S_t) = 
\{P_t(d_i \mid \Gamma(S_{MSD}), S_t) \\
\mid d_i \in D, \sum_{i=1}^{I} P_t(d_i \mid \Gamma(S_{MSD}), S_t) = 1\}
\end{multline}
where $P_t(d_i\mid \Gamma(S_{MSD}), S_t)$ \footnote{For brevity, $P_t(D \mid \Gamma(S_{MSD}), S_t)$ and $P_t(d_i\mid \Gamma(S_{MSD}), S_t)$ are referred to as $P_t(D)$ and $P_t(d_i)$.} represents the estimated probability of disease $d_i$ at iteration $t$, given the medical knowledge $\Gamma(S_{MSD})$ and the accumulated dialogue $S_t$. 
This approach enables Dr.APP to provide more transparent and informative reasoning than traditional LLMs. For example, instead of listing possible causes without prioritization—e.g., \textit{Possible causes include orthostatic hypotension, cervical spondylosis, vertigo}—Dr.APP generates a dynamic disease probability distribution, such as \textit{Orthostatic Hypotension: 0.22, Cervical Spondylosis: 0.19, Vertigo: 0.17, \dots}, which updates as the dialogue progresses.

\subsection{Question Generation}
\label{sec:question_generation}
The initial APP-user dialogue $S_1=\{q_1, r_1\}$ is often limited or imprecise, as users may use non-standard terminology or provide vague descriptions that do not directly align with clinical definitions. After estimating the disease probability $P_t(D)$, a follow-up question is needed to refine the diagnosis. At each iteration $t$, Dr.APP generates a question pool $Q_{t+1}$ guided by the MSD Manual. This grounded information is retrieved from sections such as \textit{Diagnosis} and \textit{What a doctor does}, in both the professional and consumer versions, denoted as $\Upsilon(S_{MSD})$. This ensures that the questions are both clinically reliable and symptom-specific. The set of candidate questions is represented as: $Q_{t+1} = \{q_1, \dots, q_{K}\}$, %K \leq 5$ 
where $K$ is the maximum number of questions considered per iteration. 

\subsection{One-more Step Thinking \& Conditional Probability Generation}\label{sec:one_more_think}
For each candidate question $q_{k} \in Q_{t+1}$, Dr.APP thinks one step ahead by anticipating possible patient responses. A set of plausible responses is generated by the LLM for each candidate question, given the current dialogue $S_t$. The set of responses for question $q_k$ is denoted as $R_{k} = \{r_{k}^{1}, \dots, r_{k}^{L}\}$, where $L$ is the number of generated responses. For example, the question \textit{``Can you describe what you feel when you experience dizziness?''} may yield answers such as \textit{``The room spins''}, \textit{``I feel light-headed''}, or \textit{``I lose balance but nothing spins.''}  

For each disease $d_i \in D$, the conditional probability of receiving a specific response $r_{k}^{l}$ is computed as $P(r_{k}^{l} \mid \Gamma(d_i))$, where $\Gamma(d_i)$ represents the relevant medical information for disease $d_i$ retrieved from the MSD Manual. For instance, the response \textit{``I feel light-headed''} may have a higher probability under Orthostatic Hypotension (\textit{e.g.,} 0.4), but lower probability under Vertigo (\textit{e.g.,} 0.1). 

\subsection{Question Selection} \label{sec:question_selection}
Then we use Bayesian active learning to select the optimal question from the candidate pool $Q_{t+1}$. Once responses for each candidate question are generated, Dr.APP computes the virtual next-step disease probability distribution $P(d_i|q_k)$ using Bayesian inference. The joint probability of observing both the response $r_{k}^{l}$ and the disease $d_i$ can be represented as: 
\begin{equation}
    P(r_{k}^{l}, d_i) = P(r_{k}^{l} \mid \Gamma(d_i)) \cdot P_t(d_i)
\end{equation}
Applying the law of total probability, the posterior probability of each disease $d_i$ after receiving the responses to question $q_k$ then can be updated as:
\begin{equation}
    P(d_i \mid {q_k}) = \frac{\sum_{l=1}^{L}P(r_{k}^{l}, d_i)}{\sum_{j=1}^I \sum_{l=1}^{L}P(r_{k}^{l}, d_j)}
\end{equation}

To select the optimal follow-up question $q_{t+1}^*$ for the next iteration, Dr.APP evaluates the expected entropy of each candidate question $q_k$:
\vspace{-5pt}
\begin{equation}
    H_{q_k} = -\sum_{i=1}^I P(d_i \mid q_k) \cdot logP(d_i\mid q_k)
\end{equation}
The follow-up question is then selected by minimizing entropy, ensuring that the question yields the greatest expected information gain: 
\vspace{-5pt}
\begin{equation}
    q_{t+1}^* = \arg\min_{q_{k}\in Q_{t+1}} H_{q_k}
\vspace{-5pt}
\end{equation}
After asking the optimal question $q_{t+1}^*$, the user's response $r_{t+1}$ is incorporated into the dialogue, forming $S_{t+1}$. By iteratively updating the diagnosis probability distribution $P_{t+1}(D)$ and selecting the optimal follow-up question, Dr.APP progressively reaches the final diagnosis $d^*$.

\subsection{Human-Centric Communication}
\label{sec:human_friendly}
To make the diagnostic process more accessible to users without a medical background, Dr.APP simplifies complex terms and symptom descriptions. When asking each optimal question $q_t^*$, Dr.APP is prompted to use clear, easy-to-understand language, such as \textit{``Simplify medical terminology and jargon into everyday language,''} to ensure effective communication and reduce misunderstandings. 

Individuals may not always recognize or describe abnormal behaviors or symptoms from a clinical perspective. To address this, Dr.APP provides contextual hints and is explicitly prompted to frame questions in simple yes/no or multiple-choice formats. For example, instead of asking a broad question like \textit{``Have you eaten anything unusual?''}, it might ask \textit{``Have you consumed foods like milk or beverages like soda (e.g., Coke)?''}

Even with simplified yes/no questions, users may still struggle with vague symptom descriptions or unfamiliar medical terminology. Dr.APP addresses this by using descriptive examples. Rather than asking \textit{``Do you feel dizzy?''}, Dr.APP might ask: \textit{``Are you experiencing a feeling of losing balance, or does it seem like your surroundings are spinning or moving, even when everything is still?''} This helps users express their symptoms more accurately.

\section{Experiment}
\subsection{Dataset}
To evaluate our proposed approach, Dr.APP, we use a subset of the ReMeDi dataset \cite{yan2022remedi}, which contains real-world multi-turn conversations between doctors and patients. This ensures that the dialogues reflect natural clinical interactions with realistic variability. We use ReMeDi-base, which originally consists of 1,557 labeled dialogues. In this dataset, doctors' responses are annotated with seven action types: \textit{``Informing'', ``Inquiring'', ``Chitchat'', ``QA'', ``Recommendation'', ``Diagnosis'', and ``Others''}. We extract 329 real-world conversations that exclusively contain the \textit{``Diagnosis''} label and randomly select 100 for our study. These cases cover 72 distinct diseases across 18 specialties, including Otolaryngology (\textit{e.g.,} Allergic Rhinitis), Gynecology (\textit{e.g.,} Polycystic Ovary Syndrome), and Gastroenterology (\textit{e.g.,} Gastroesophageal Reflux Disease).

\subsection{Patient Simulator}
To evaluate different LLM-simulated medical assistants, we simulate realistic patients based on selected ReMeDi dialogues. For each case, we reconstruct a patient profile using DeepSeek-v3, summarizing structured characteristics such as symptoms, age, intention and personality. These attributes are compiled into a comprehensive patient persona used to guide a simulated patient agent, which interacts with a target LLM-simulated medical assistant (e.g., DeepSeek-v3, GPT-4o). The simulated patient agent responds using only the constructed persona, without access to the original ReMeDi dialogue. To better mimic realistic patient behavior, the simulated patient agent is further guided with instructions such as: \textit{``Reasonably incorporate daily life details that align with the patient’s personality and background.''} This setup forms the foundation of a benchmark for evaluating diagnostic capabilities through natural, multi-turn interactions. By combining real-world personas with consistent and adaptive patient responses, our benchmark closely replicates real-world consultations in a controlled and reproducible environment.

%-------------------------------------------------------------------------
\begin{figure*}[hbt!]
    \centering
    \includegraphics[width=0.98\linewidth]{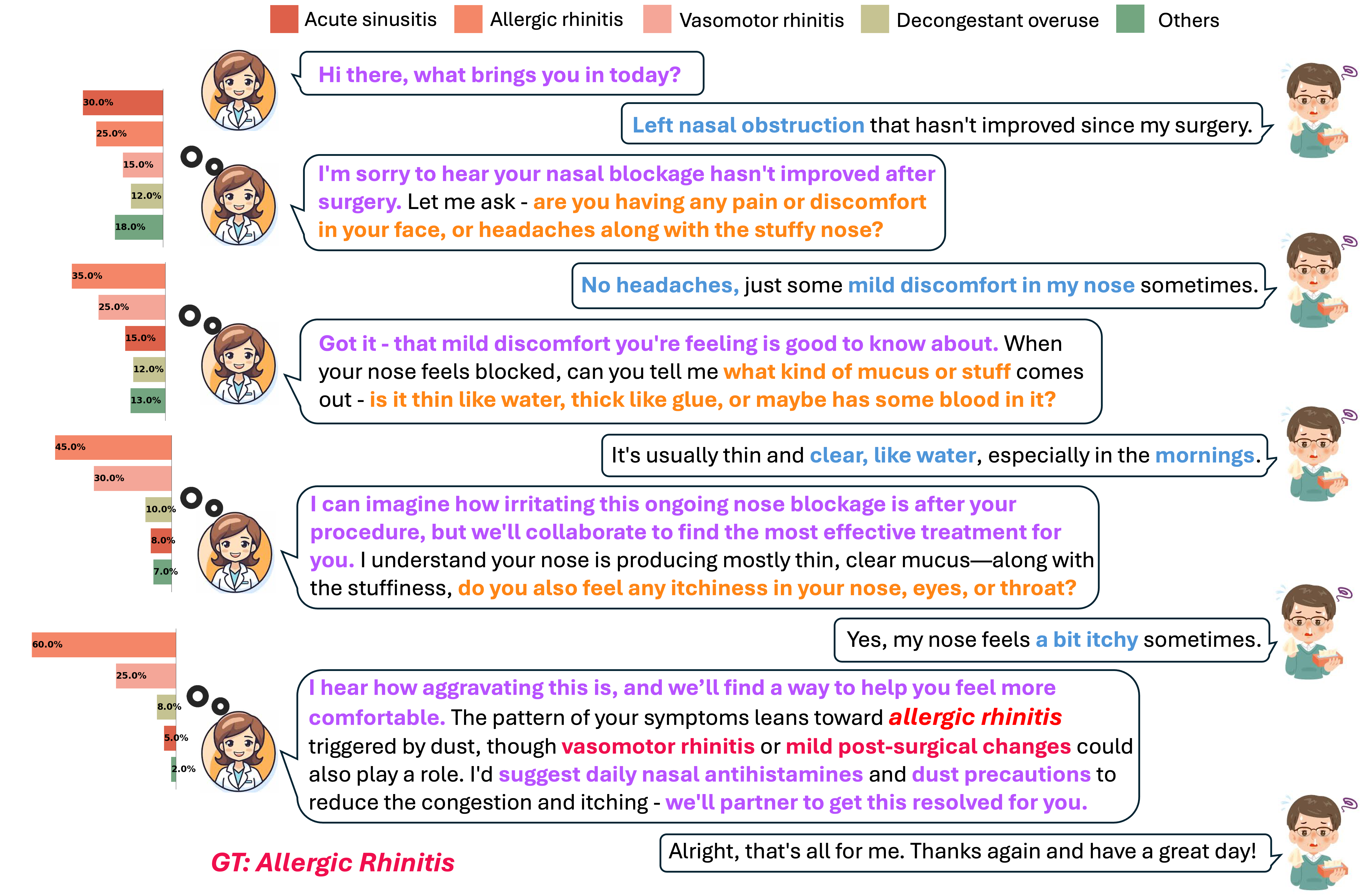} %width=0.75\linewidth
    \caption{An APP case study of human-centric multi-turn dialogue based on medical guidelines. The estimated disease distribution is updated with the progression of the conversation. Disease items from top to bottom in the last iteration: Allergic Rhinitis, Acute Sinusitis, Vasomotor Rhinitis, Decongestant Overuse and others. The ground truth is Allergic Rhinitis, where our diagnosis is Allergic Rhinitis. \textcolor[RGB]{78, 149, 217}{Blue} represents user-described symptoms, \textcolor[RGB]{255, 133, 21}{Orange} indicates questions raised by APP, \textcolor[RGB]{240, 14, 77}{Red} highlights the diagnosis, and \textcolor[RGB]{184, 82, 255}{Purple} shows human-centric features.}
    \label{fig:case_study}
\end{figure*}
%-------------------------------------------------------------------------

\subsection{Experimental Setup}
In our experimental setup, we set the number of candidate questions per iteration to $K=5$. For each question, at least two and at most of five plausible patient responses ($2 \leq L \leq 5$) are generated.

\subsection{Evaluation Matrix}
\subsubsection{Accuracy}
We invited three experts, each with over five years of biomedical experience, to evaluate the quality of the predicted diagnoses. Each expert independently rated the predictions using a 5-point scale adapted from \cite{kanjee2023accuracy}, reflecting alignment with the ground truth. A score of 0 indicates no relevance, while a score of 5 denotes an exact match (see Appendix \ref{supp:survey_doctor}). This evaluation provides a nuanced assessment of diagnostic performance. Experts also assessed the model’s trustworthiness based on its alignment with grounded guidelines, its ability to gather relevant patient information, the clarify of explanations, and its support for better doctor-patient relationship.

\subsubsection{Entropy}
Given the current disease probability distribution $P_t(D)$, the goal is to increase diagnostic confidence and rule out unlikely conditions through multi-turn dialogue. We use entropy as a quantitative measure to assess diagnostic confidence and interpretability \footnote{Figure \ref{fig:overflow}(a)(b) present potential diseases without indicating their likelihood, while Figure \ref{fig:case_study} shows how Dr.APP distinguishes between more and less probable diseases.}. The entropy at iteration $t$ is calculated as: $H_t = - \sum_{i=1}^{I}P_t(d_i)\cdot logP_t(d_i)$, where $P_t(d_i)$ is the probability of disease $d_i$ and $I$ is the total number of possible diseases at iteration $t$. A reduction in entropy over successive dialogue turns indicates increased diagnostic confidence. 

\begin{table*}[]
\centering
\renewcommand{\arraystretch}{1.2}
\caption{\textbf{Diagnosis Accuracy (5-point score) Comparison with SOTA Methods}: This table presents the diagnostic accuracy of three common specialties and the overall performance across all 18 specialties. APP-DeepSeek-v3 achieves the highest overall accuracy in both one-shot and multi-turn evaluations, demonstrating the effectiveness of multi-turn interactions driven by statistical modeling and grounded medical guidelines.}
\resizebox{0.9\linewidth}{!}{
\begin{tabular}{?c?cccc?cccc?}
\specialrule{1.5pt}{0pt}{0pt}
\multirow{2}{*}{Model} & \multicolumn{4}{c?}{One-Shot}                     & \multicolumn{4}{c?}{Multiple-Turn}               \\
\cdashline{2-9}
                       & Neurology & Cardiology & Nephrology & Overall & Neurology & Cardiology & Nephrology & Overall \\
\specialrule{1.5pt}{0pt}{0pt}

LLaMA-70B               & 3.04      & 1.72   & 1.13            & 2.77   & 3.19      & 2.11   & 1.66            & 2.93   \\
Claude-3               & 2.90      & 1.94   & 1.73            & 2.94   & 3.23      & 2.44   & 2.20            & 2.98   \\
GPT-4o                 & 2.81      & 2.27   & 1.93            & 2.90   & 3.10      & 2.72   & 2.06            & 2.96   \\
\rowcolor{red!5}
QWen-72B              & 2.47      & 2.27   & 1.53            & 2.89   & 2.86      & 2.66   & 1.93            & 3.07   \\
\rowcolor{red!5}
APP-QWen-72B          & 3.09      & \textbf{2.61}   & 2.13            & 3.06   & 3.43      & 2.83   & 2.53            & 3.14   \\
\rowcolor{blue!5}
DeepSeek-v3            & 3.00    & \textbf{2.61}  & 1.93            & 3.08   & 3.28      & 2.83   & 2.13            & 3.17   \\
\rowcolor{blue!5}
APP-DeepSeek-v3        & \textbf{3.33}      & \textbf{2.61}   & \textbf{2.20}            & \textbf{3.35}  & \textbf{3.90}      & \textbf{3.00}   & \textbf{2.86}            & \textbf{3.59}  \\
\specialrule{1.5pt}{0pt}{0pt}
\end{tabular}} \label{table:accuracy}
\end{table*}
\subsubsection{Human-Centric}
We invited five participants without medical backgrounds to evaluate the human-centric qualities of Dr.APP using recorded consultation transcripts. Twenty case studies were randomly selected, each containing five versions: Claude-3, GPT-4o, DeepSeek-v3, APP-DeepSeek-v3, and the original real-world dialogue from ReMeDi. Each participant independently reviewed all 100 dialogues (five per case), with the model order randomized for each case to ensure fair comparison. Participants rated each dialogue across four aspects: the \textit{accessibility score} captured how clear and easy the language was for non-medical users; the \textit{empathy score} reflected the degree of empathetic communication shown during the conversation; the \textit{relevant response rate} evaluated whether the model appropriately responded to follow-up questions before moving on; and the \textit{relationship fostering score} measured the model’s ability to build a supportive and trusting consultation. All scores were rated on a scale from 1 to 5, as detailed in Appendix \ref{supp:patient_question}.

\subsection{Accuracy Analysis versus Baselines}
To evaluate the diagnostic accuracy of Dr.APP, we compared its performance against SOTA LLMs across multiple medical domains, including neurology, cardiology, and nephrology (Table \ref{table:accuracy}). The ``Overall'' column represents the diagnostic performance averaged across all 18 clinical specialties. We conducted evaluations in both single-turn and multi-turn diagnostic settings, with the multi-turn setup involving six rounds of iterative questioning to refine their diagnoses.

In the one-shot setting, where models generated diagnoses based only on the initial user input without follow-up interaction, APP-DeepSeek-v3 achieved the highest overall accuracy of 3.35 on the 5-point scale. It outperformed all other models, including DeepSeek-v3 (3.08), Claude-3 (2.94) and GPT-4o (2.90). Notably, APP-QWen-72B also showed improved performance over its base model QWen-72B, achieving an overall score of 3.06 compared to QWen-72B’s 2.89. In the multi-turn setting, APP-DeepSeek-v3 again outperformed other methods, reaching an overall accuracy of 3.59, with particularly strong results in neurology (3.90). Similarly, APP-QWen-72B benefited significantly from the multi-turn interaction, improving to 3.14, outperforming its base model QWen-72B (3.07).

%-------------------------------------------------------------------------
\begin{figure}[ht]
            \centering
            \vspace{-5pt}\includegraphics[scale=0.3]{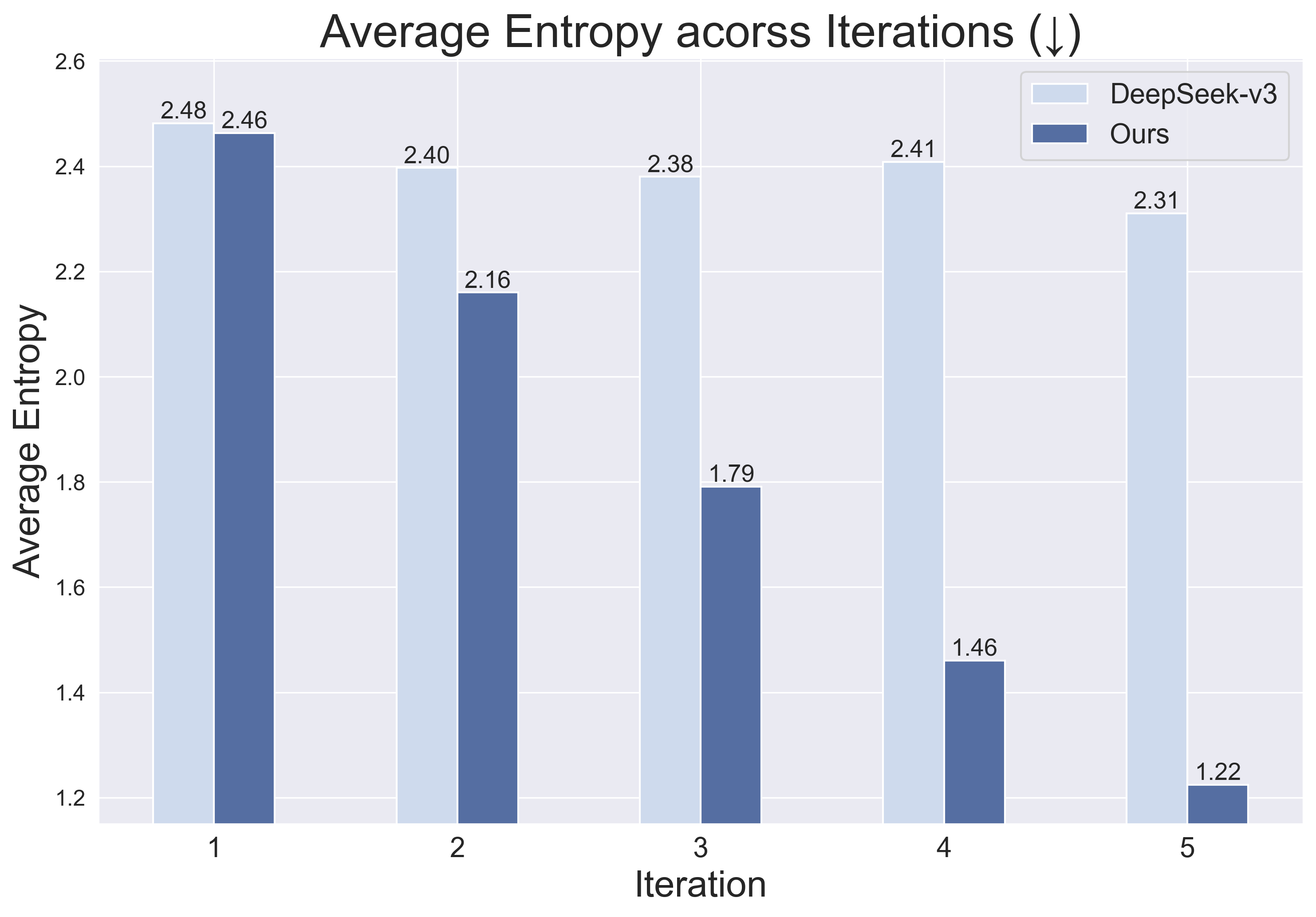}
            \vspace{-10pt}
    \caption{\textbf{Entropy Comparison across Iterations.} Dr.APP consistently shows a sharper decrease in entropy, indicating increased diagnostic confidence and reduced uncertainty through iterative dialogues.}
            \label{fig:entropy}
\vspace{-15pt}
\end{figure}
%-------------------------------------------------------------------------

%-------------------------------------------------------------------------
\begin{figure}[ht]
            \centering
    \includegraphics[scale=0.2]{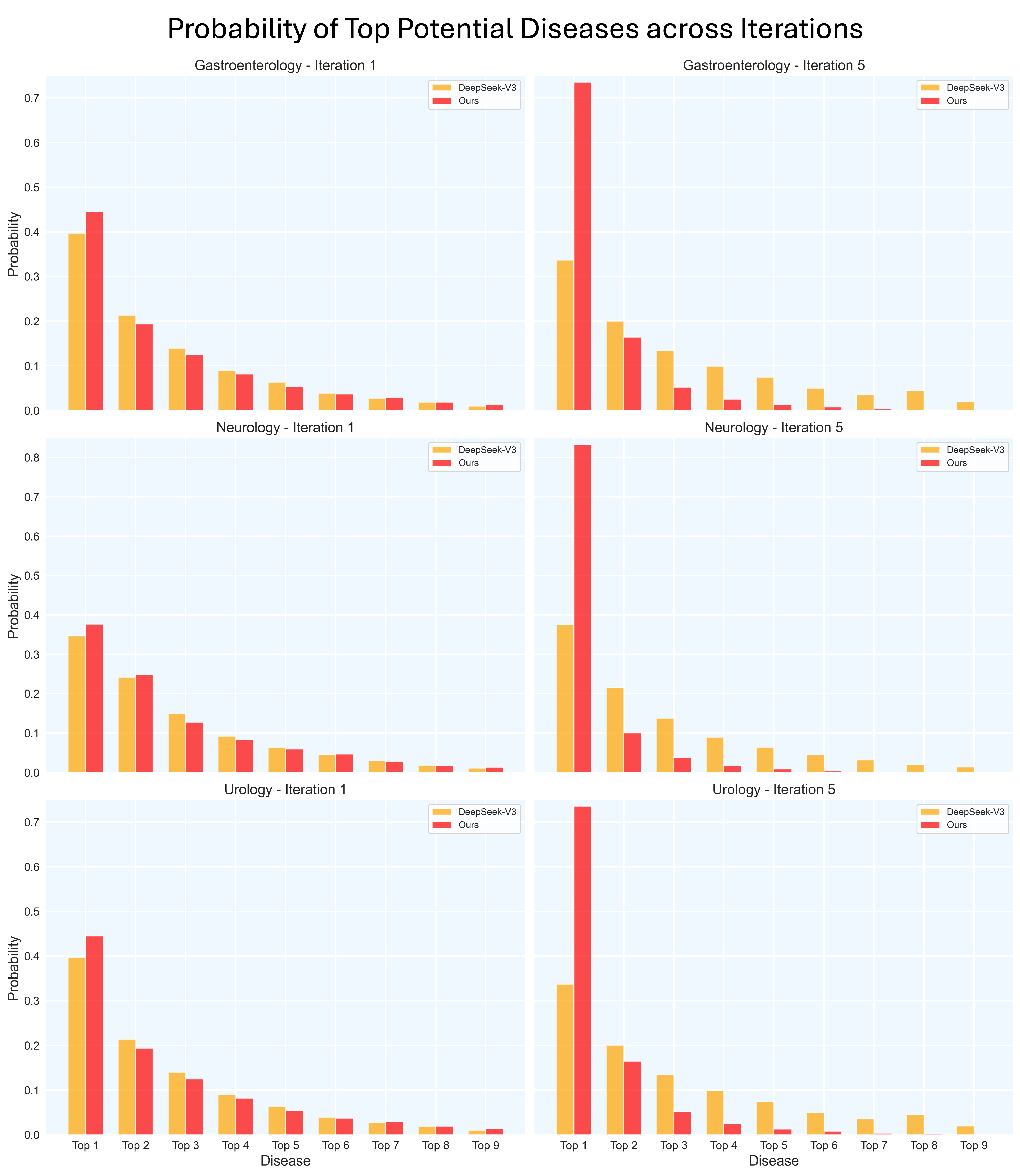}
            \vspace{-10pt}     \caption{\textbf{Confidence Analysis across Iterations.} APP-DeepSeek-v3 shows increased confidence in the top predicted disease while reducing confidence in less likely conditions over multiple iterations, demonstrating improved diagnostic confidence with interpretability.}
    \label{fig:disease_dist}
\vspace{-10pt}
\end{figure}
%-------------------------------------------------------------------------

\subsection{Confidence Analysis across Iterations}
Figure \ref{fig:entropy} illustrates the evolution of diagnostic confidence by comparing entropy values between APP-DeepSeek-v3 and the DeepSeek-v3 baseline across iterations. In the initial step, before any follow-up questions, Dr.APP exhibits slightly lower diagnostic uncertainty than DeepSeek-v3 (entropy of 2.46 \textit{versus} 2.48).
%, likely due to its reliance on verified medical sources for initial reasoning.
As iterations progress, Dr.APP shows a sharper and more consistent decline in entropy, refining diagnoses more effectively. After five iterations, Dr.APP reduces its entropy to 1.22, indicating high confidence in its predictions, whereas the baseline remains at 2.31, suggesting persistent uncertainty. This reduction highlights Dr.APP’s superiority in reducing diagnostic uncertainty and improving prediction confidence.

Figure \ref{fig:disease_dist} further shows how the distribution of top potential diseases evolves across iterations for different specialties, including gastroenterology, neurology, and otolaryngology. The results indicate that Dr.APP consistently assigns higher confidence to the most probable disease while reducing confidence in less likely conditions. This leads to a clearer separation in probability rankings. The widening gap suggests that Dr.APP systematically refines its predictions, improving diagnostic clarity and reducing ambiguity over multiple interactions.

By presenting intermediate reasoning and confidence adjustments across iterations, Dr.APP improves model transparency and diagnostic certainty. The increasing confidence reduces ambiguity, enabling more reliable and trustworthy medical guidance. These improvements ultimately foster greater user trust in AI-assisted diagnosis while enhancing clinical reliability and usability.

\begin{table}[ht]
\vspace{-5pt}
\centering
\renewcommand{\arraystretch}{1.2}
\caption{\textbf{Human-Centric Metric Comparison.} The table reports the average score on a 5-point scale. APP-DeepSeek-v3 consistently outperforms baseline models and the real-world doctor-patient dialogue, demonstrating improved alignment with human-centric features.}
\vspace{-5pt}
\resizebox{0.48\textwidth}{!}{
\centering
\begin{tabular}{?c?cccc?}
\specialrule{1.5pt}{0pt}{0pt}
\multirow{2}{*}{Methods} & \multicolumn{4}{c?}{Human-Centric Metric} \\
\cdashline{2-5}
                         & Accessibility    & Empathy    & Relevant Response    & Foster Relation    \\
\hline
Calude-3                & 3.18      & 2.54         & 3.78          & 2.96                       \\
GPT-4o      & 2.86       & 2.34        & 3.38        & 2.56               \\
DeepSeek-v3 & 3.10         & 2.42          & 3.50        & 2.56                \\
Real-world Dialogue & 3.48        & 2.78       & 3.24        & 2.92                 \\
\rowcolor{blue!5}
APP-DeepSeek-v3        & \textbf{4.54} & \textbf{4.60} & \textbf{4.48} & \textbf{4.54}     \\
\specialrule{1.5pt}{0pt}{0pt}
\end{tabular}}\label{table:human_centric}
\vspace{-5pt}
\end{table}

\subsection{Human-Centric Analysis}
% with Real-world Dialogue
Our human-centric system, Dr.APP, shows notable improvements across four key dimensions: user accessibility, question empathy, response relevance, and relationship fostering. As shown in Table \ref{table:human_centric}, Dr.APP consistently outperforms both baseline models and real-world dialogues from ReMeDi.

Specifically, in terms of accessibility, Dr.APP achieved an average score of 4.54, significantly surpassing real-world dialogues (3.48) and all baseline models. This suggests the system effectively presents medical information in a more user-friendly manner. For empathy, Dr.APP scored 4.60, markedly higher than the original dialogues (2.78), highlighting its ability to generate more compassionate and supportive responses, which may help reduce user anxiety and improve the consultation experience. In response relevance, Dr.APP reached 4.48, outperforming the second-best model, Claude-3, by a margin of 0.7. Finally, in fostering relational engagement, Dr.APP achieved a score of 4.54, indicating a stronger ability to build trust and rapport with users. Overall, these results demonstrate that Dr.APP substantially enhances the human-centric quality of dialogue, contributing to improved user understanding, satisfaction, and engagement.

\section{Related Work}
\textbf{Medical Dialogue with LLMs.} Multi-turn conversational LLMs are crucial for healthcare, as they can iteratively gather and interpret relevant patient information, enabling more accurate and context-aware decision-making than single-turn systems \cite{li2025beyond}. Most existing studies focus on improving response accuracy in medical dialogues through supervised fine-tuning on large-scale healthcare datasets \cite{chen2023bianque, bao2023disc, pieri2024bimedix}. Some methods aim to enhance the model's information-seeking ability through multi-turn interactions \cite{zhang2023huatuogpt, li2024mediq, li2025aligning}. A smaller subset incorporates grounding mechanisms using external medical knowledge sources during inference to improve reliability \cite{yang2024zhongjing}, while others prioritize human-centric evaluation through patient-oriented metrics that simulate real-world clinical engagement \cite{yang2024zhongjing, tu2025towards}.

\section{Conclusion}
In this study, we introduce Dr.APP, the first human-centric, LLM-based medical assistant for grounded reasoning and transparent diagnosis. Dr.APP enhances diagnostic accuracy and reliability by integrating authoritative medical guidelines and leveraging Bayesian active learning to optimize follow-up questioning. Through entropy minimization, Dr.APP improves transparency and progressively increases diagnostic confidence. To support evaluation in realistic settings, we introduce a new benchmark that simulates multi-turn medical consultations, using patient agents constructed from real-world doctor–patient dialogues. Our experiments demonstrate that Dr.APP significantly outperforms both one-shot and current multi-turn LLM baselines in diagnostic accuracy. Entropy analysis confirms that Dr.APP rapidly reduces diagnostic uncertainty over successive iterations, leading to greater confidence in its predictions. Human evaluation further shows that Dr.APP prioritizes accessibility and empathetic communication, responds with more medically relevant information, and fosters a stronger patient–doctor relationship. By bridging clinical expertise and human-centric dialogue, Dr.APP promotes greater user trust and engagement.

\section*{Limitations}
Despite its advancements, Dr.APP has several limitations that warrant further exploration. 

First, while Dr.APP reduces diagnostic uncertainty through entropy minimization at each step, it may converge to a local minimum rather than achieving the global minimum. This limitation arises because Dr.APP selects the next question based on immediate entropy reduction, rather than considering the long-term impact of each question on overall diagnostic certainty. As a result, suboptimal question sequences may occasionally lead to delayed or less efficient diagnosis. To address this, future work could explore reinforcement learning-based optimization or multi-step planning strategies that anticipate future interactions rather than relying solely on greedy entropy reduction. Additionally, incorporating global uncertainty estimation techniques, such as Bayesian optimization or Monte Carlo dropout methods, could further enhance robustness in question selection and diagnostic confidence.

Second, while Dr.APP integrates medical guidelines to improve diagnostic reliability, but it may still be constrained by the quality and coverage of these guidelines. In this paper, we use the MSD Manual as an example of grounded medical knowledge, but many other real-world medical sources exist. Expanding the system to incorporate additional medical knowledge bases could further enhance its clinical applicability.

Finally, most of our evaluation relies on simulated patient interactions and human assessments of recorded consultation transcripts. However, real-world clinical trials are needed to validate Dr.APP’s effectiveness in actual medical settings. Future research should focus on deploying Dr.APP in real-world consultations and assessing its impact on patient outcomes, physician workload, and healthcare accessibility.

\section*{Acknowledge}
Ms. Zhu is supported by the Engineering and Physical Sciences Research Council (EPSRC) under grant EP/S024093/1 and Global Health R\&D of the healthcare business of Merck KGaA, Darmstadt, Germany, Ares Trading S.A. (an affiliate of Merck KGaA, Darmstadt, Germany), Eysins, Switzerland (Crossref Funder ID: 10.13039/100009945). Mr. Liu is supported by the Clarendon Fund. Mr. Wu is supported by the Engineering and Physical Sciences Research Council (EPSRC) under grant EP/S024093/1 and GE HealthCare. 


%\newpage
\bibliography{custom}

\newpage
\appendix
\section{Appendix}

%%%%%%%%%%%%%%%%%%%%%%%%%%%%%%%%
\subsection{Survey Question - Doctor}
\label{supp:survey_doctor}
Thank you for participating in this survey. Please assess each response generated by the model based on the following criteria. Provide your rating on a scale from 1 to 5, where 1 is the lowest and 5 is the highest. You may also leave optional comments to clarify your reasoning.
\begin{enumerate}
    \item \textbf{Diagnosis Accuracy}
    \begin{itemize}
    \item How accurate is the model’s predicted diagnosis compared to the actual diagnosis?
    \item Rating Scale: \textbf{0:} Completely unrelated to the actual diagnosis. \textbf{2:} Related, but unlikely to be useful. \textbf{3:} Closely related – may still be helpful. \textbf{4:} Very close – minor difference but clinically similar. \textbf{5:} Exact match with the ground truth diagnosis. (Note: there is no score of 1.)
    \item \textbf{Optional Comment}: Please explain your score or provide examples where the model matched or missed the diagnosis.
    \end{itemize}
    
    \item \textbf{Reliability Score (Rel.)}
    \begin{itemize}
        \item Does the model’s predicted disease align with verified medical knowledge?
        \item Rating Scale: \textbf{1:} Completely incorrect - contradicts medical guidelines. \textbf{2:} Mostly incorrect - with major inaccuracies. \textbf{3:} Partially correct - but has some errors. \textbf{4:} Mostly accurate - only minor inconsistencies. \textbf{5:} Fully accurate - aligns with established medical knowledge.
        \item \textbf{Optional Comment}: Do you notice any inaccuracies or missing medical reasoning?
    \end{itemize}

        \item \textbf{Fostering the Relationship (FR)}
    \begin{itemize}
        \item How would you rate the model’s behavior in fostering a relationship with the patient?
        \item Rating Scale: \textbf{1:} Very poor – no rapport or engagement. \textbf{2:} Poor – minimal effort to build trust. \textbf{3:} Fair – some acknowledgment but limited warmth. \textbf{4:} Good – shows care and encourages connection. \textbf{5:} Excellent – empathetic, respectful, and partnership-oriented.
        \item \textbf{Optional Comment}: Did the model help build trust, connection, or respect? Please provide examples.
    \end{itemize}

    \item \textbf{Gathering Information (GI)}
    \begin{itemize}
        \item How would you rate the model’s ability to gather relevant information from the patient?
        \item Rating Scale: \textbf{1:} Very poor – fails to gather necessary details. \textbf{2:} Poor – asks limited or irrelevant questions. \textbf{3:} Fair – gathers some useful information. \textbf{4:} Good – asks mostly appropriate and clear questions. \textbf{5:} Excellent – thoroughly elicits meaningful and context-aware input.
        \item \textbf{Optional Comment}: Did the model miss any critical details or show strong information-gathering behavior?
    \end{itemize}

    \item \textbf{Providing Information (PI)}
    \begin{itemize}
        \item How would you rate the model’s ability to provide understandable and accurate information to the patient?
        \item Rating Scale: \textbf{1:} Very poor – unclear or incorrect information. \textbf{2:} Poor – hard to follow or overly technical. \textbf{3:} Fair – mostly understandable but lacks clarity. \textbf{4:} Good – clear with some complexity. \textbf{5:} Excellent – clear, accessible, and well-structured.
        \item \textbf{Optional Comment}: Did the model communicate effectively and support patient understanding?
    \end{itemize}
    
\end{enumerate}

%%%%%%%%%%%%%%%%%%%%%%%%%%%%%%%%
\subsection{Survey Question - Patient}
\label{supp:patient_question}
Thank you for participating in this survey. Please assess each response generated by the model based on the following criteria. Provide your rating on a scale from 1 to 5, where 1 is the lowest and 5 is the highest. You may also leave optional comments to clarify your reasoning.

\begin{enumerate}
    \item \textbf{Accessibility Score (Acc.)}
    \begin{itemize}
        \item How easy is it for you to understand the question posed by the model?
        \item Rating Scale: 
        \textbf{1:} Very difficult - full of medical jargon. \textbf{2:} Mostly difficult - require effort to interpret. \textbf{3:} Somewhat clear - but have some medical terms that may be confusing. \textbf{4:} Mostly clear - only minor terminology issues.
        \textbf{5:} Completely clear - no unnecessary medical jargon.
        \item \textbf{Optional Comment}: Are there any terms or phrases that made it hard to understand? Could you provide examples?
    \end{itemize}

    \item \textbf{Empathy Score (Emp.)}
    \begin{itemize}
        \item How empathetic does the model feel to you during the conversation?
        \item Rating Scale: \textbf{1:} Completely robotic - no sense of empathy. \textbf{2:} Somewhat cold - little acknowledgment of concerns.
        \textbf{3:} Neutral - acknowledges concerns but lacks warmth. \textbf{4:} Shows care and reassurance - with some empathetic responses. \textbf{5:} Very empathetic - makes you feel understood and supported.
        \item \textbf{Optional Comment}: Is there anything that felt particularly empathetic or lacking in care?
    \end{itemize}

    \item \textbf{Relevant Response Rate (RRR)}
    \begin{itemize}
        \item Does the model directly answer your follow-up questions before moving on?
        \item Rating Scale: \textbf{1:} Completely ignores the question or gives an irrelevant response. \textbf{2:} Partially answers - but lacks detail. \textbf{3:} Answers the question - but may miss key points. \textbf{4:} Mostly relevant - only minor gaps. \textbf{5:} Fully relevant -directly answers with the right level of detail.

        \item \textbf{Optional Comment}: Are there any responses that felt off-topic or incomplete?
    \end{itemize}
    
    \item \textbf{Fostering the Relationship (FR)}
    \begin{itemize}
        \item How would you rate the model’s behavior in fostering a relationship during the interaction?
        \item Rating Scale: \textbf{1:} Very poor – no rapport, closed-off. \textbf{2:} Poor – limited openness or empathy. \textbf{3:} Fair – acknowledges patient but lacks warmth. \textbf{4:} Good – shows care and builds some trust. \textbf{5:} Excellent – builds connection, respect, and partnership.
        \item \textbf{Optional Comment}: Did the model make you feel acknowledged, respected, or supported? Please share examples.
    \end{itemize}
\end{enumerate}

\end{document}